\documentclass[runningheads]{llncs}

\usepackage[T1]{fontenc}
\usepackage{graphicx}
\usepackage{color}
\usepackage{balance}
\usepackage[table,xcdraw,x11names]{xcolor}
\usepackage{multirow}
\usepackage{xspace}
\usepackage{enumitem}
\usepackage{amssymb}
\usepackage{wrapfig}
\usepackage{ulem}
\usepackage{subcaption}

\begin{document}
\title{AI-Generated Images: What Humans and Machines See When They Look at the Same Image}
\titlerunning{AI-Generated Images: Human versus Machines}
%
\author{Silvia Poletti,	Justin Ilyes, Marcel Hasenbalg, David Fischinger, Martin Boyer}

\authorrunning{S. Poletti et al.}
%
\institute{Austrian Institute of Technology, Vienna, Austria}
%
%
\maketitle              
\begin{abstract}
The misuse of generative AI in online disinformation campaigns highlights the urgent need for transparent and explainable detection systems. In this work, we investigate how detectors for AI-generated images can be more effective in providing human-understandable explanations for their predictions. To this end, we develop a suite of detectors with various architectures and fine-tuning strategies, trained on our large-scale photorealistic fake image dataset, AIText2Image, and assess their performance on state-of-the-art text-to-image AI generators. We integrate 16 different explainable AI (XAI) methods into our detection framework, and the visual explanations are comprehensively refined and evaluated through a novel approach that prioritizes human understanding of AI-generated images, using both textual and visual responses collected from a survey of 100 participants. This framework offers insights into visual-language cues in fake image detection and into the clarity of XAI methods from a human perspective, measuring the alignment of XAI outputs with human preferences.

\end{abstract}
%
%
%


\begin{figure*}[t]
	\centering
	\includegraphics[width=0.99\linewidth, trim=0 5.8cm 0.4cm 5.8cm, clip]{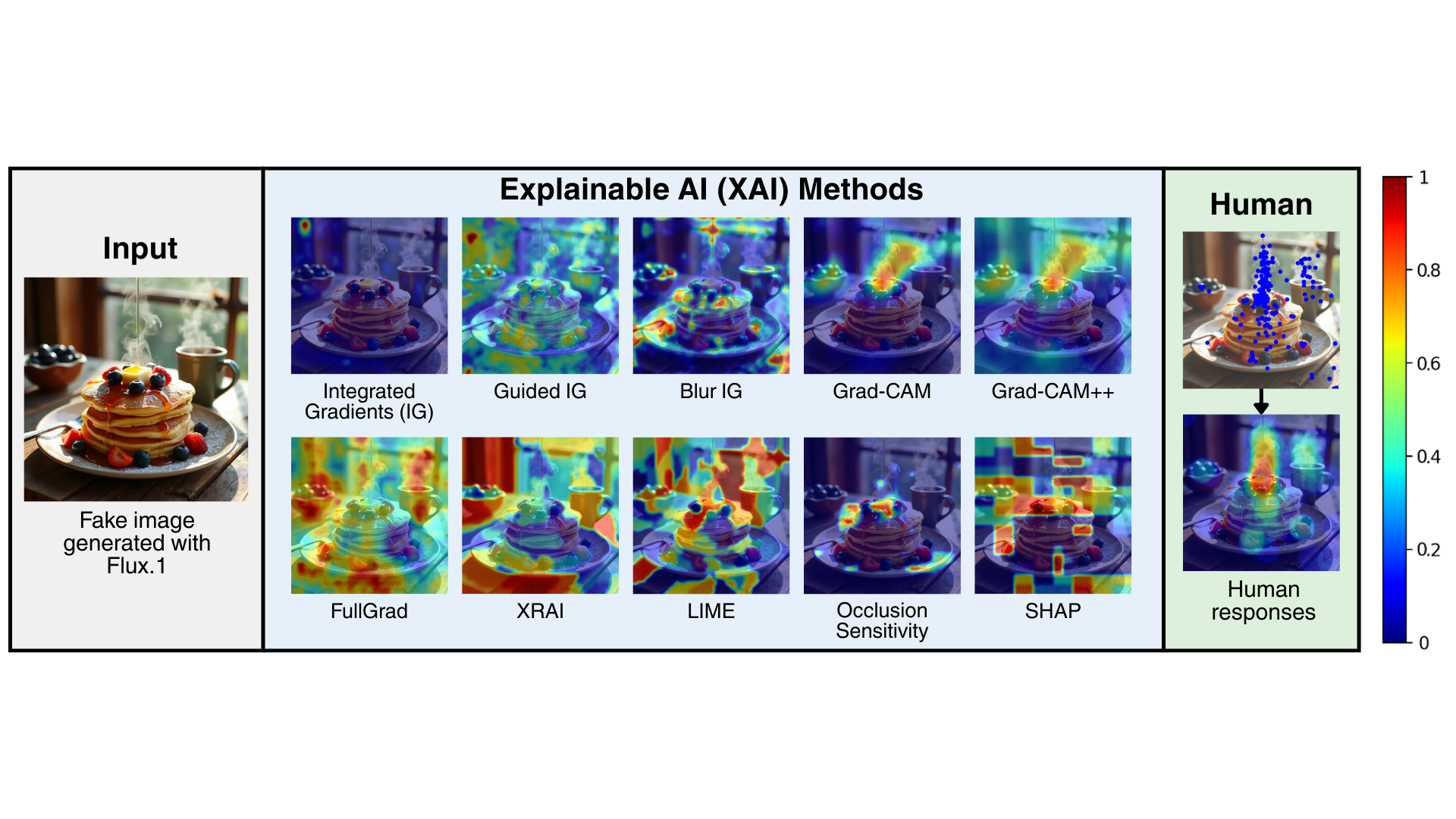}
	\caption{Overview of visual explanations of the XAI methods when applied to the same AI detection model, and the visual explanation based on the survey's human responses.
		To reduce redundancy, XAI methods  yielding visually comparable results are omitted.}
	\label{fig:masks}
\end{figure*}


\section{Introduction}
\label{sec:intro}

Photographs were once trusted as reliable evidence due to the technical difficulty of realistic manipulation,
but today generative AI models allow to easily create fake images directly from a simple text prompt. 
The misuse of generative AI can influence democratic elections through fake news 
and defame individuals with deepfakes, causing significant societal disruptions. 
Notably, 
humans show limited ability to detect image tampering \cite{nightingale2017can}, and
governments worldwide 
are currently developing laws and regulations to mitigate the negative impact of synthetic media on society, for example with the AI Act by EU.
However, given the widespread accessibility of generative AI, legal frameworks alone cannot prevent its harmful use, and technological safeguards such as AI-generated image detectors (hereafter referred to as AI detectors) must be developed in parallel.
Since detectors typically output only a confidence score as prediction, they offer limited transparency of the decision-making process. Explainable AI (XAI) is then essential to clarify how and why a model classifies an image as fake. 

To address the under-explored area of explainability in AI-generated image detection, this study makes the following contributions.
We introduce AIText2Image, a large-scale dataset of photorealistic AI-generated images from modern text-to-image AI generators, and train detection models with multiple fine-tuning strategies, achieving robust accuracy in detecting outputs from modern text-to-image generators.
We integrate 16 XAI methods into our framework, proposing a novel categorization based on the methods' visual features and prioritizing human preferences in their evaluation. To this aim, we investigate the alignment between image regions highlighted in the visual explanations and those indicated by 100 human participants in a user study. Importantly, participants are not merely asked to rate given explanations, but to provide their own. An illustrative example is in Figure~\ref{fig:masks}. Finally, we further analyze textual explanations in the context of multi-modal detection and explainability.

\section{Related work}
\label{sec:related}

As AI image generation advances, discerning real from synthetic images is proven to grow increasingly difficult \cite{ha_007163_2024}.
According to Lu~et al.~\cite{human-survey}, humans can correctly classify only 61\% of images as real or synthetic,
while state-of-the-art AI detectors largely surpass this performance, and therefore should serve as a tool to support human judgment. 
While prior studies have largely focused on humans’ ability to detect AI-generated images, our work is the first to rigorously investigate through a user study how humans interpret and understand model predictions and reasoning, using widely adopted XAI techniques.

Model reasoning in fake image classification concerns different cues. Semantic-based detectors focus on physical inconsistencies or abnormalities \cite{borji2023qualitative}, while spatial-based detectors consider the image domain and examine edges, textures or gradients \cite{zhong_007101_2024,lorenz_007183_2023,arai_005000_2024}.
Alternative approaches shift from the image domain and classify vector embeddings \cite{ojha_004000_2023,sha_003000_2023} or expose the AI generator-specific fingerprints in the image's frequency domain \cite{wang_002000_2023,ma_007184_2023,bammey2023synthbuster}, especially for detecting Generative Adversarial Networks (GANs) \cite{zhu_001000_2024,wolter_007179_2022}.
This work focuses on human-understandable complex cues, thus mainly semantic and spatial, and does not consider AI generators leaving clearly detectable fingerprints, as this could bias the AI detectors to focus only on these low-level artifacts.

Bird~and~Lotfi~\cite{bird2024cifake} address the black box problem using the XAI method Grad-CAM \cite{gradcam},
showing that only local patches of the image are considered to predict synthetic images. Our study further investigates which local image's areas are relevant for AI detection and extends beyond a single type of XAI method by integrating 16 different techniques and further analyzes the overlap between the obtained saliency maps and human expectations.
Notably, research on human-AI collaboration has recently gained traction. Liu~et al.~\cite{human-attention} introduced HAG-XAI, a human attention–guided method that uses eye-tracking data to align saliency maps with human perception.
However, fine-tuning well-established, pre-trained image classification models remains the overall simplest, while effective approach for detecting AI-generated images, as more complex architectures typically provide only limited improvements in accuracy and generalization ~\cite{zhu_001000_2024}. 
In particular, Zhu~et al.~\cite{zhu_001000_2024} and Dogoulis~et al.~\cite{dogoulis2023improving} demonstrated ResNet-50 and CLIP-ViT backbones can achieve state-of-the-art results for the task.

\section{Detection of AI-generated images}
\label{sec:detector}

We examine two key aspects of explainability in AI detection models. First, \textbf{model architecture}: we assess explainability of convolutional layers in Convolutional Neural Networks (CNN) and self-attention layers in Visual Transformers (ViTs), using widely adopted backbone models to minimize the impact of specific model design choices and derive more universal XAI properties. Second, \textbf{fine-tuning strategy}: we start from models pre-trained on ImageNet \cite{imagenet} and implement four transfer learning approaches, each differing in the number of model's retrained layers.

\textbf{CNN backbone.} We consider ResNet-50 \cite{resnet} and distinguish ResNet50[fc] in which only the last fully connected layer is retrained, ResNet50[fc+l4] in which the fourth CNN block is additionally retrained, and then ResNet50[fc+l4+l3] in which all layers from the third CNN block onward are retrained.

\textbf{ViT backbone.} We use ViT-B-16 \cite{vit} and retrain all model's parameters.

A total of 162k images from Microsoft COCO were used as natural samples during training, depicting objects in a realistic, cluttered environment and offering more complex scenarios than other datasets primarily intended for image classification. 
The \textbf{AIText2Image} training dataset includes 209k fake images from modern text-to-image AI generators that are among the most accessible, well-established, and widely used. Images are generated with engineered prompts covering a combination of subjects, locations, objects, image attributes and backgrounds.
The generators are reported in Table~\ref{tab:eval} under the Train cluster, and the diversity of their output quality is essential to train detectors to be robust enough to variations of generative models. 

\renewcommand{\arraystretch}{1.1}
\begin{center}
	\begin{table*}[h!]
		\caption{F1-scores of detection models across 13 image subsets. The number of images in each subset and the type of data (\textcolor{blue}{R} for real and \textcolor{red}{F} for fake) is reported. All AI detectors are trained under the same early-stopping regime.
		}
		\vspace{0.2cm}
		\centering
		\scriptsize
		\begin{tabular}{
				>{\centering\arraybackslash}p{1.5cm}|
				>{\centering\arraybackslash}p{0.573cm}
				>{\centering\arraybackslash}p{0.573cm}
				>{\centering\arraybackslash}p{0.573cm}
				>{\centering\arraybackslash}p{0.573cm}
				>{\centering\arraybackslash}p{0.573cm}|
				>{\centering\arraybackslash}p{0.573cm}
				>{\centering\arraybackslash}p{0.573cm}
				>{\centering\arraybackslash}p{0.573cm}
				>{\centering\arraybackslash}p{0.573cm}
				>{\centering\arraybackslash}p{0.573cm}
				>{\centering\arraybackslash}p{0.573cm}
				>{\centering\arraybackslash}p{0.573cm}
				>{\centering\arraybackslash}p{0.573cm}|
				>{\centering\arraybackslash}p{0.4cm}}
			\cline{2-14}
			\rule{0pt}{85pt} & \rotatebox{90}{Microsoft COCO\cite{coco}} & \rotatebox{90}{Midjourney 5.2 \cite{midjourney5}} & \rotatebox{90}{SDXL 0.9 \cite{sdxl}} & \rotatebox{90}{SDXL 1.0 \cite{sdxl}} & \rotatebox{90}{DALL-E 2 \cite{dalle2}} & \rotatebox{90}{ImageNet100 \cite{imagenet}} & \rotatebox{90}{Pascal-VOC2012 \cite{pascal-voc-2012}} & \rotatebox{90}{OpenImages-v7 \cite{openimg}} & \rotatebox{90}{Flux.1 \cite{flux}} & \rotatebox{90}{Adobe Firefly \cite{firefly}} & \rotatebox{90}{Midjourney 5\&6 \cite{midjourney5,midjourney6}} & \rotatebox{90}{DALL-E 3 \cite{dalle3}} & \rotatebox{90}{GLIDE \cite{glide}} &\\ 
			\cline{2-14}
			& \multicolumn{5}{c|}{\textbf{Train}} & \multicolumn{8}{c|}{\textbf{Evaluation}} & \\ 
			& $162$k & $132$k & $17$k & $52$k & $9$k & $10$k & $10$k & $10$k & $6.5$k & $1.1$k & $7.5$k & $1.2$k  & 1k&\\ 
			& \textcolor{blue}{R} & \textcolor{red}{F} & \textcolor{red}{F} & \textcolor{red}{F} & \textcolor{red}{F} & \textcolor{blue}{R} & \textcolor{blue}{R} & \textcolor{blue}{R} & \textcolor{red}{F} & \textcolor{red}{F} & \textcolor{red}{F} & \textcolor{red}{F} & \textcolor{red}{F}&\\ 
			\hline
			\multicolumn{1}{|c|}{ \cellcolor{gray!20} \textbf{AI detectors}}& \multicolumn{13}{c|}{\cellcolor{gray!20} \textbf{F1-score}}&\multicolumn{1}{c|}{\cellcolor{gray!20} \textbf{Avg.}}\\
			\hline
			\multicolumn{1}{|c|}{ResNet50[fc]} & 97.2 & 97.3 & 99.5 & 99.4 & 89.1 & 87.1 & 82.5 & 96.4  & 98.2 & 92.2 & \textbf{96.7} & \textbf{98.1} & 84.9 & \multicolumn{1}{c|}{93.8} \\
			\hline
			\multicolumn{1}{|c|}{ResNet50[fc+l4]} & 99.7 & \textbf{99.8} & 99.8 & \textbf{100} & 97.1 & 97.0 & 92.4 & 99.4 & \textbf{98.8} & \textbf{96.4} & 87.4 & 92.1 & 86.2 & \multicolumn{1}{c|}{95.9} \\
			\hline
			\multicolumn{1}{|l|}{ResNet50[fc+l4+l3]} & \textbf{99.8} & 99.4 & 99.5 & 99.8 & 93.7 & \textbf{99.2} & \textbf{96.6} & \textbf{99.6} & 95.8 & 79.0 & 89.5 & 96.2 & 74.8 & \multicolumn{1}{c|}{94.1} \\
			\hline
			\multicolumn{1}{|c|}{ViT-B-16} & 99.2 & 99.6 & \textbf{99.9} & 99.9 & \textbf{98.2} & 94.2 & 89.6 & 97.5  & 94.9 & 86.9 & 93.8 & 97.0 & \textbf{89.9} & \multicolumn{1}{c|}{94.7} \\
			\hline
		\end{tabular}
		\label{tab:eval}
	\end{table*}
\end{center}


\section{Model explainability}
\label{sec:detection}

We focus on 16 explainable AI (XAI) methods for image classification, which provide local visual explanations by highlighting parts of the input image that are relevant for the classifier's prediction. 
If an image correctly classified as fake, these explanations reveal the visual cues that the AI detector associates with synthetic content.
We only consider post-hoc methods 
that preserve the classifier's performance and avoid model retraining~\cite{ante-hoc}. 

Given an AI detector and an RGB input image $\mathbf{I}\in[0,255]^{w \times h \times 3}$, an XAI method produces an output matrix $\mathbf{M}\in\mathbb{R}^{w \times h}$ in which higher values indicate greater importance of the corresponding input pixels for the AI detector's prediction.
$\mathbf{M}'\in[0,1]^{w \times h}$ is the normalized matrix and is referred to as XAI mask. Depending on the specific XAI method, the normalization requires one or more of the following operations: k-Means clustering, Gaussian kernel smoothing with Gaussian blur, Min-Max Scaling and Percentile scaling.

\begin{description}
	
	\item[Min-Max Scaling:] 
	\begin{equation}
		\label{eq:min-max}
		\mathbf{M}' = \frac{\mathbf{M} - \min(\mathbf{M})}{\max(\mathbf{M}) - \min(\mathbf{M})}.
	\end{equation}
	
	\item[Percentile Scaling:] 
	\begin{equation}
		\label{eq:perc}
		M_{ij}' = p_{ij} \quad such\;that \quad M_{ij} = Q_{p_{ij}}(\mathbf{M}) \quad\forall 1\leq i\leq w, \;1\leq j\leq h
	\end{equation}
	where $0 \leq p_{ij} \leq 100$ and $Q_p$ is the $p$-th percentile operation.
\end{description}

\begin{figure*}[h!]
	\centering
	\includegraphics[width=0.8\linewidth, trim=0 0 17cm 0, clip]{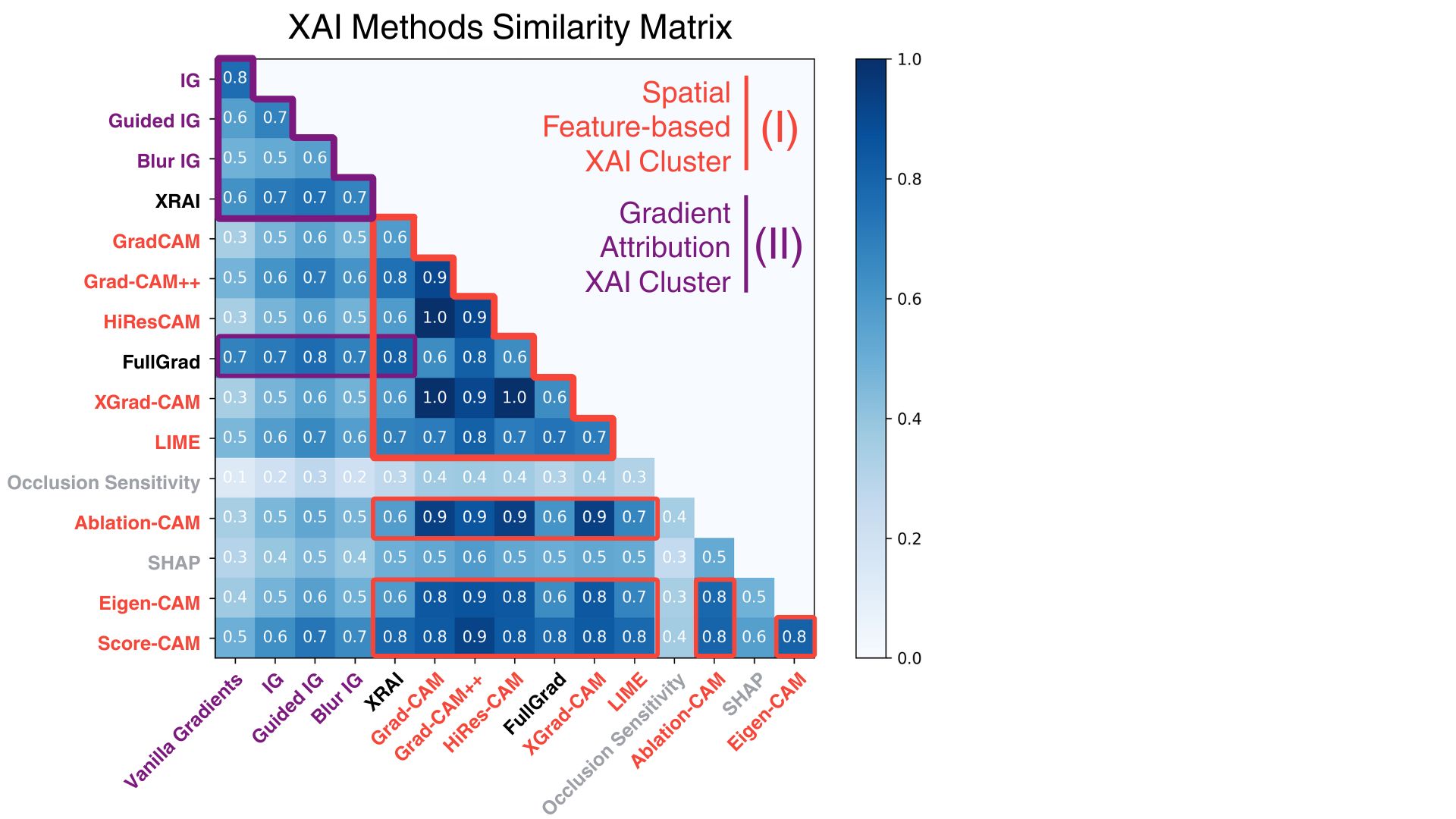}
	\caption{XAI methods comparison based on the cosine similarity scores between pairs of XAI masks generated by different methods applied to the same AI detector, ResNet50[fc+l4]. Results for other detectors 
		are comparable. The plot shows average scores computed over 100 images from AIText2Image, evenly distributed across the AI generators.}
	\label{fig:intra-sim}
\end{figure*}

Liu~et al.~\cite{human-attention} categorize XAI methods solely based on their functioning, distinguishing between gradient-based methods using the classifier's gradients to compute feature importance, and perturbation-based methods evaluating changes in the model's output by altering the input features. Instead, we propose a novel categorization based on visual intra-similarity, indicating common discriminative features in the image domain:

\begin{enumerate}[label=\textbf{\Roman*.}]
	\item \textbf{Spatial Feature-based XAI Cluster}\label{cluster:spatial}: Class Activation Mapping (CAM) XAI methods, namely Grad-CAM~\cite{gradcam}, Grad-CAM++~\cite{gradcam++}, HiResCAM~\cite{hirescam}, XGrad-CAM~\cite{xgradcam}, Ablation-CAM~\cite{ablcam}, Eigen-CAM~\cite{eigencam}, Score-CAM~\cite{scorecam} and FullGrad~\cite{fullgrad}, exhibit scores consistently above or close to 0.8. LIME~\cite{lime} also show a good similarity with CAM methods, since both capture spatial patterns: CAM captures them from the model's layers while LIME from the input image via segmentation.
	\item \textbf{Gradient Attribution XAI Cluster}: Gradient-based XAI methods such as Vanilla Gradients~\cite{vanillag1}, Integrated Gradients~\cite{ig} (IG) and its variants, Guided IG~\cite{guidedig} and Blur IG~\cite{blurig}, have significant similarity. FullGrad is also similar to the three IG methods, as it is designed to capture gradient-based contributions across the network. 
\end{enumerate}

XRAI~\cite{xrai} belongs to both clusters, as it builds upon IG while also integrating spatial information through a gradient-based attention mechanism.
In contrast, SHAP~\cite{shap} and Occlusion Sensitivity~\cite{os} (OS) produce distinctly different XAI masks from the other methods, one featuring rectangular regions and the other sparse, scattered activation regions (see Figure~\ref{fig:masks}).


\section{Human study}
\label{sec:survey}
Human perception of artifacts in AI-generated images is under-explored in literature. To address this, we use responses from a human survey to answer two key questions: \textit{What visual cues do humans most associate with AI generation? How consistent are these with popular XAI Methods?}
We submitted a survey to $100$ candidates, required to be between $18$ and $50$ years old, and to be familiar with the internet and social media. Participants were shown fake images and asked to provide 1) a \textbf{visual response}, by indicating one or two points in the image they believed had traces of AI generation, and 2) a \textbf{textual response}, by writing a textual explanation for their choice.
The survey instructions clarified that AI generation traces can be anything perceived as artificial, from textures or visual inconsistencies to elements that seem out of place in the image scene context. 
The survey includes $52$ fake images from DALL-E 2, SDXL 1.0, Midjourney 5\&6, and Flux.1, presenting diverse quality in terms of visual artifacts.

The experiments employ an image labeling system. 
Label \textbf{CONTEXT} (\textit{Context is informative}) is assigned when AI generation can be inferred from contextual cues, such as floating objects defying gravity. The absence of this label implies that the image can be identified as AI-generated based solely on visual artifacts. Label \textbf{ENV} (\textit{Visible environment}) indicates a discernible background, while its absence suggests an anonymous or heavily blurred background. Other labels are \textbf{ANIMAL} (\textit{Visible animals}), \textbf{HUMAN} (\textit{Human subject}), \textbf{HANDS} (\textit{Visible hands}), \textbf{VIP} (\textit{VIP subject}) assigned to images depicting recognizable individuals, \textbf{SOLO} (\textit{Single human subject}), and \textbf{PORTRAIT} (\textit{Human portrait}). 

\begin{figure*}[h!]
	\centering
	\includegraphics[width=0.99\linewidth, trim=0 12.5cm 0 0, clip]{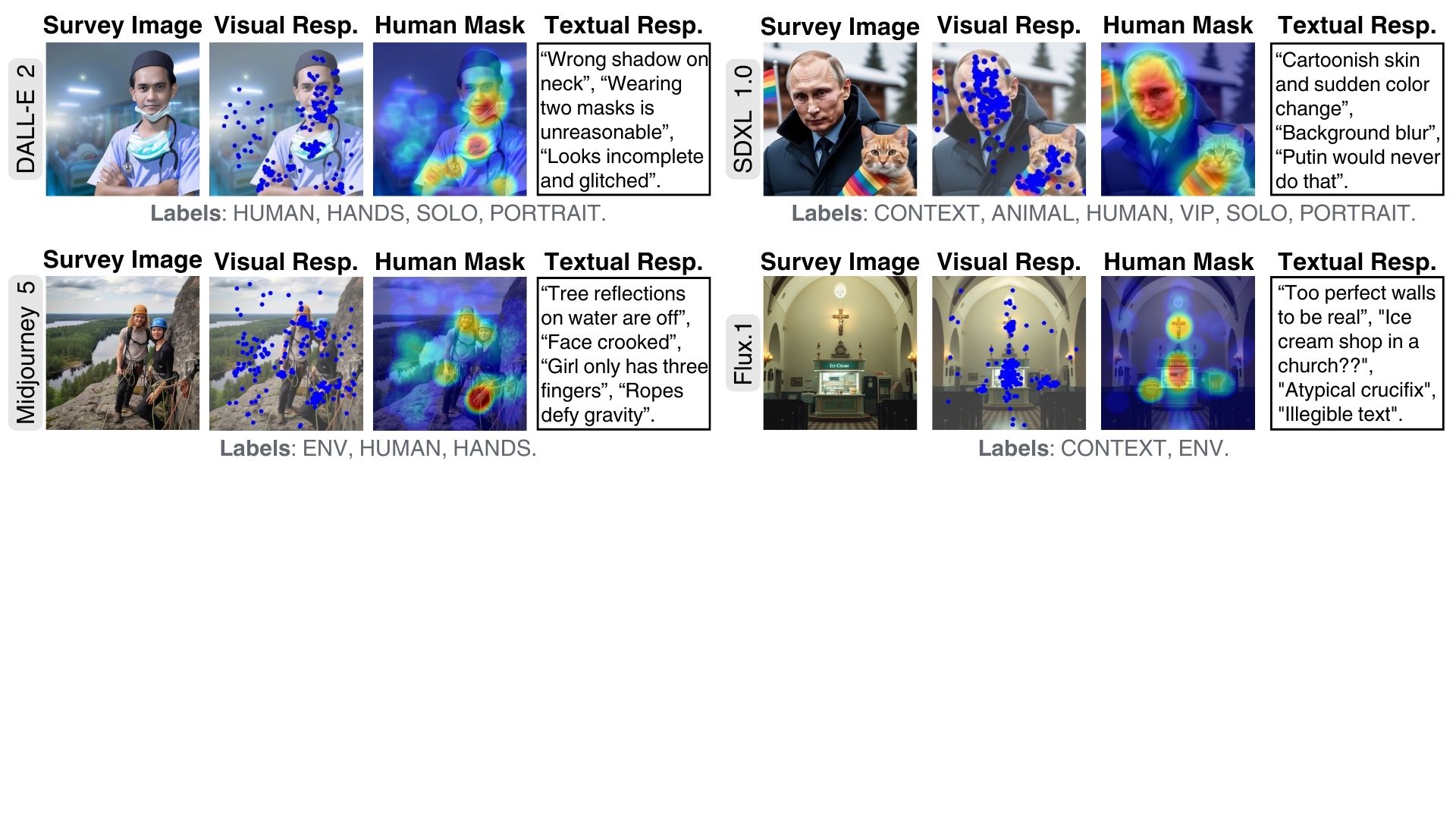}
	\caption{Survey image examples from four different generators along with the participants' visual responses and corresponding human masks and assigned image labels. Due to space limitations, only a few textual responses are shown.}
\label{fig:survey-img}
\end{figure*}

\subsection{Analysis of survey visual responses}\label{sec:visual}
\begin{figure}[h!]
	\centering
	\includegraphics[width=0.9\linewidth, trim=0 0cm 15cm 0cm, clip]{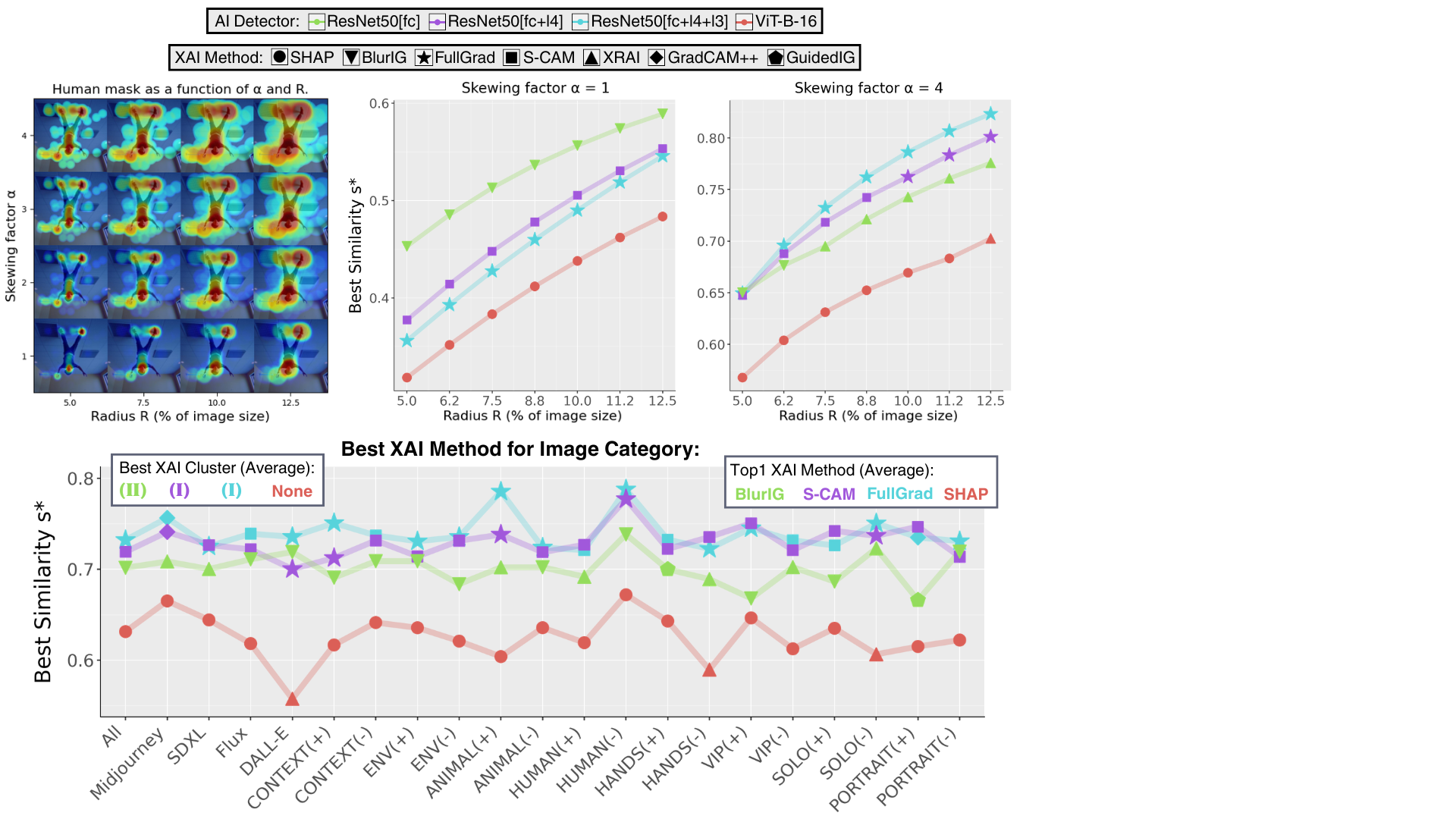}
	\caption{First row: XAI methods achieving average optimal similarity $s^*$ across all the survey images for different AI detectors, evaluated for varying values of $R$ and $\alpha$ in the human mask. Second row: Best XAI methods by image category, for a fixed $R$ equal to $8.8\%$ of the image size, and $\alpha=3$, which are the values providing less bias towards the preference of one XAI methods to another. }
	\label{fig:human_r_alpha}
\end{figure}

For each survey image, a so-called human mask $\mathbf{H}\in[0,1]^{w \times h}$ is computed, as a result of the contributions of all the visual responses (i.e.~points selected in the image) given by the participants for a picture $\mathbf{I}\in[0,255]^{w \times h \times 3}$.
The mask is generated by increasing the matrix value by a constant factor $c>0$ in a circular area of radius $R>0$ centered around each point. Then, the mask is normalized and the values are further skewed of a factor $\alpha>0$ using exponential scaling. Examples of human masks are reported in Figure~\ref{fig:survey-img} and Figure~\ref{fig:human_r_alpha}. 

Each XAI method $m_k$ provides a mask $\mathbf{M}_k\in[0,1]^{w \times h}$ of the input, representing the importance attributed to the image regions by the detector. 
The best-performing method $m_{k^*}$ should produce an XAI mask having highest similarity to the human mask, therefore providing the most human-understandable visual explanation. The faithfulness of the XAI methods, that is how accurately the explanations reflect the AI detector’s decision-making process, has already been thoroughly established in the original publications of the methods. On the other hand, in this work our focus is specifically on human-centered plausibility — how well an explanation aligns with what humans perceive as meaningful. 

Considering the cosine similarity $\mathcal{S}$, the optimal index $k^*$ and the highest similarity score $s^*$ are defined as:
\begin{equation}
	k^* =  \arg\max_{k} \mathcal{S}(\mathbf{H}, \mathbf{M}_k), \quad s^*=\max_k\mathcal{S}(\mathbf{H}, \mathbf{M}_k).
\end{equation}


As shown in Figure~\ref{fig:human_r_alpha}, the AI detectors clearly demonstrate varying compatibility with XAI methods. The highest similarity occurs in images without human subjects (\textbf{HUMAN(-)}), as they tend to absorb human attention, leading to human masks overall different from XAI masks.
High similarity values are also obtained for Midjourney pictures, which present the highest level of realism in the dataset. In contrast, DALL-E 2 images lead to an overall lower similarity, indicating that for lower-quality outputs, as those produced by DALL-E 2, the most noticeable artifacts perceived by humans differ substantially from those detected by the models.

\begin{figure}[h!]
	\centering
	
	\begin{minipage}{0.45\textwidth}
		\centering
		\includegraphics[width=\linewidth]{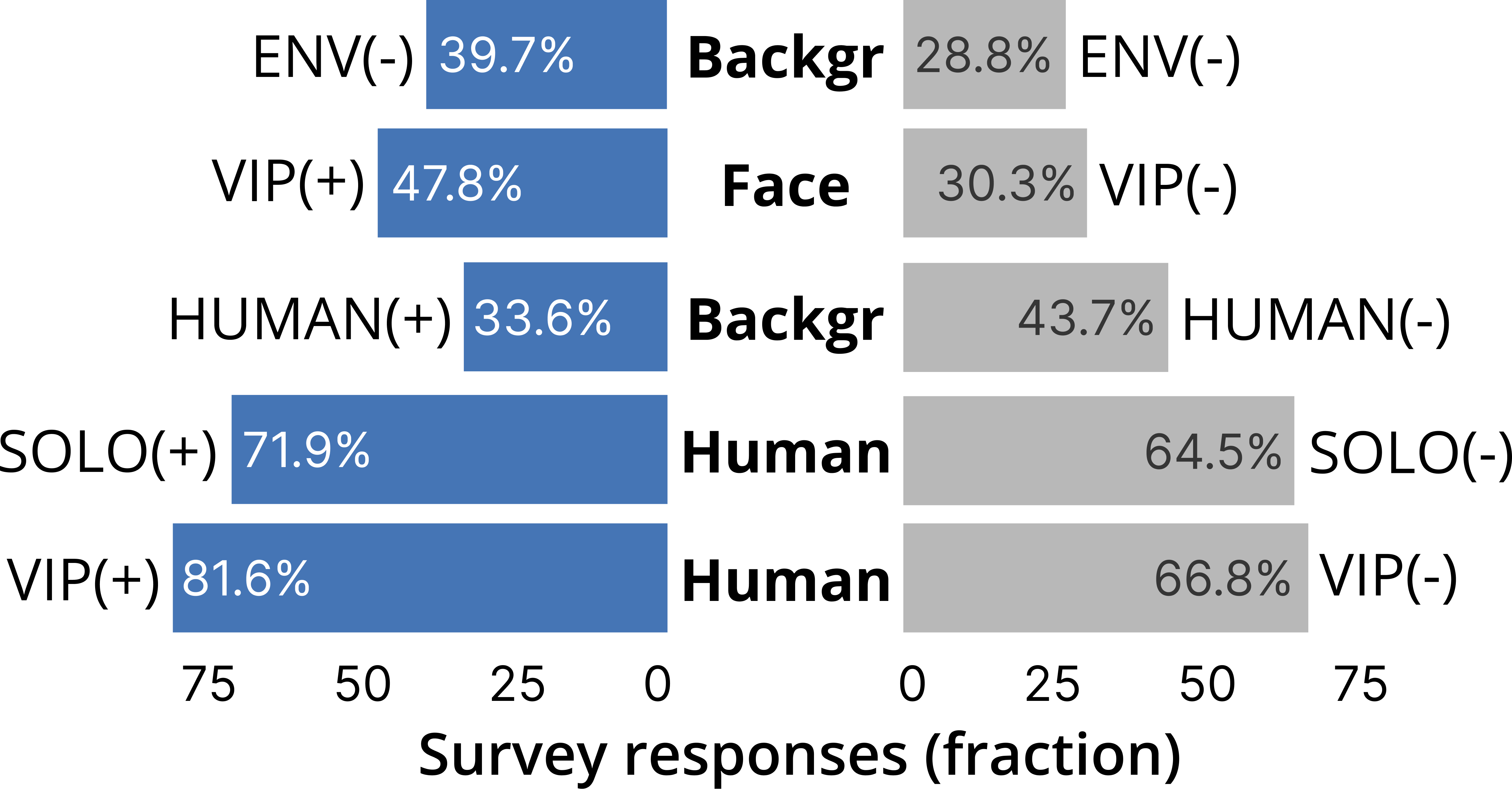}
		\\ (a)
	\end{minipage}
	\hspace{0.02\textwidth}
	\begin{minipage}{0.48\textwidth}
		\centering
		\includegraphics[width=\linewidth]{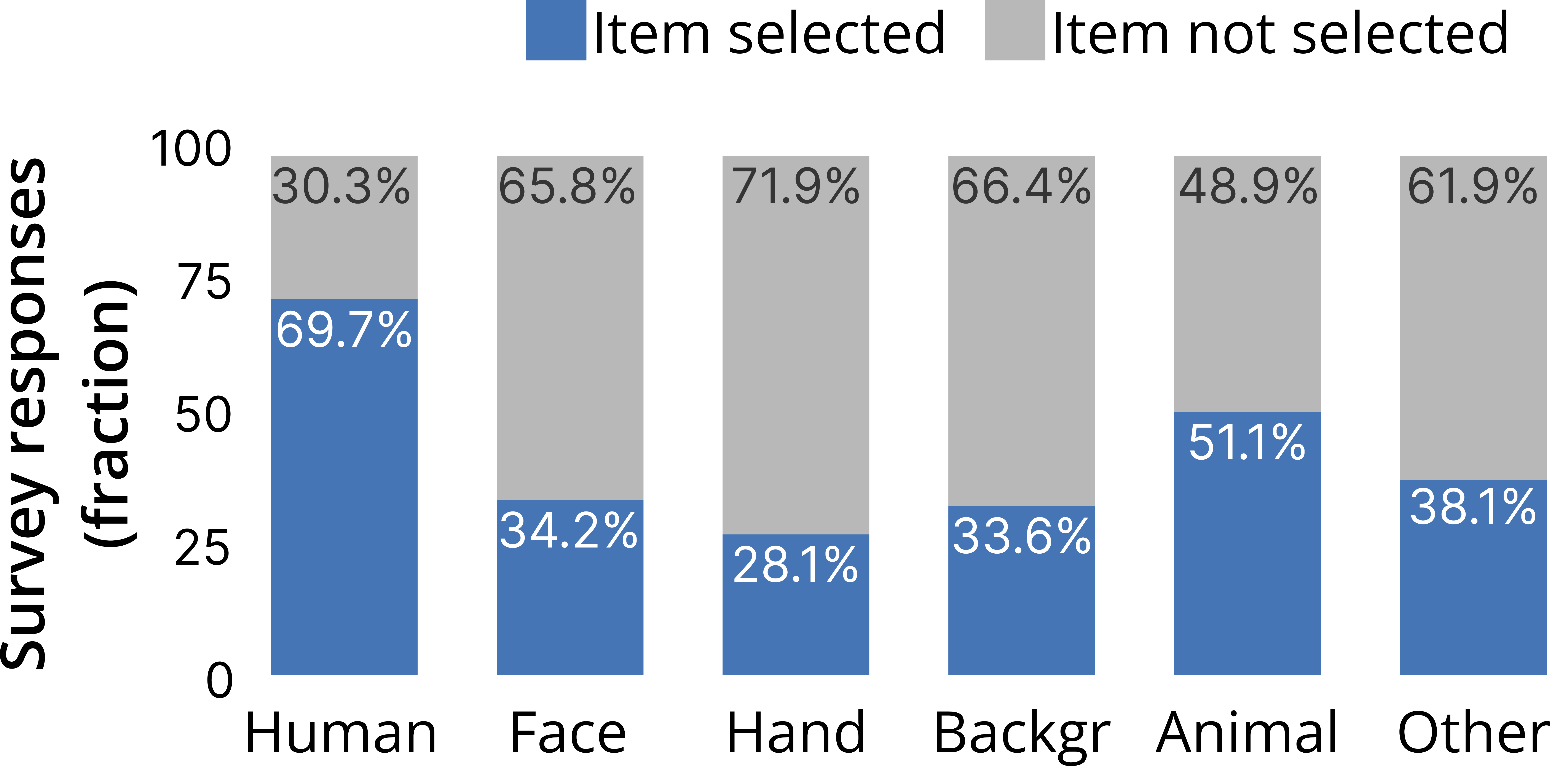}
		\\ (b)
	\end{minipage}
	
	\caption{(a) Each row represents a subset of images containing a specific item (Human, Face, or Background). Horizontal bars indicate the fraction of human responses selecting the corresponding item as indicator of AI traces, comparing two opposite image categories, e.g.~\textbf{SOLO(+)} vs. \textbf{SOLO(-)}. (b) Only images with human subjects are considered. Blue bars indicate the human responses selecting a specific item, when present in the image. Note that each response is made of two clicks on the image, potentially indicating more than one item.}
	
	\label{fig:barchart}
\end{figure}

Figure~\ref{fig:barchart} reports the changes in the fraction of survey responses where certain items were selected. For example, the human attention on the background decreases (i.e.~the fraction of selected points in the background decreases) as it gets anonymous or heavily blurred (\textbf{ENV(-)}) or when a human subject is present (\textbf{HUMAN(+)}). The presence of a famous person (\textbf{VIP(+)}) moves the focus towards the human subject and, specifically, to their face. This happens also when there is only one human subject (\textbf{SOLO(+)}) present in the picture.
When a human subject is selected ($69.7\%$ of the times), face and hands are the primary focus points, as they account for overall $62.3\%$ of the selected items.

\subsection{Analysis of survey textual responses}
\label{sec:text-h}

There are two categories of survey textual responses: those referring to the image's visual quality and those reasoning about the realism of the image's content. 
The first includes the responses of people noticing (i) errors in the perspective, lights, and reflections; (ii) visual inconsistencies, as incomplete shapes or sudden color changes; (iii) errors in the texture, resolution, color and details; (iv) unnatural blurred background; (v) distortions and visual blend of multiple items. The second category is related to (vi) errors in shape, appearance and anatomy; (vii)~lacking of imperfections; (viii) absence of physics; (ix) atypical content; (x) items presented in a novel, implausible design; (xi) atypical behavior for well-known people; (xii) presence of non-sense writing and illegible characters.

\begin{figure}[h!]
	\centering
	\includegraphics[width=0.99\linewidth, trim=0 11.5cm 0cm 0cm, clip]{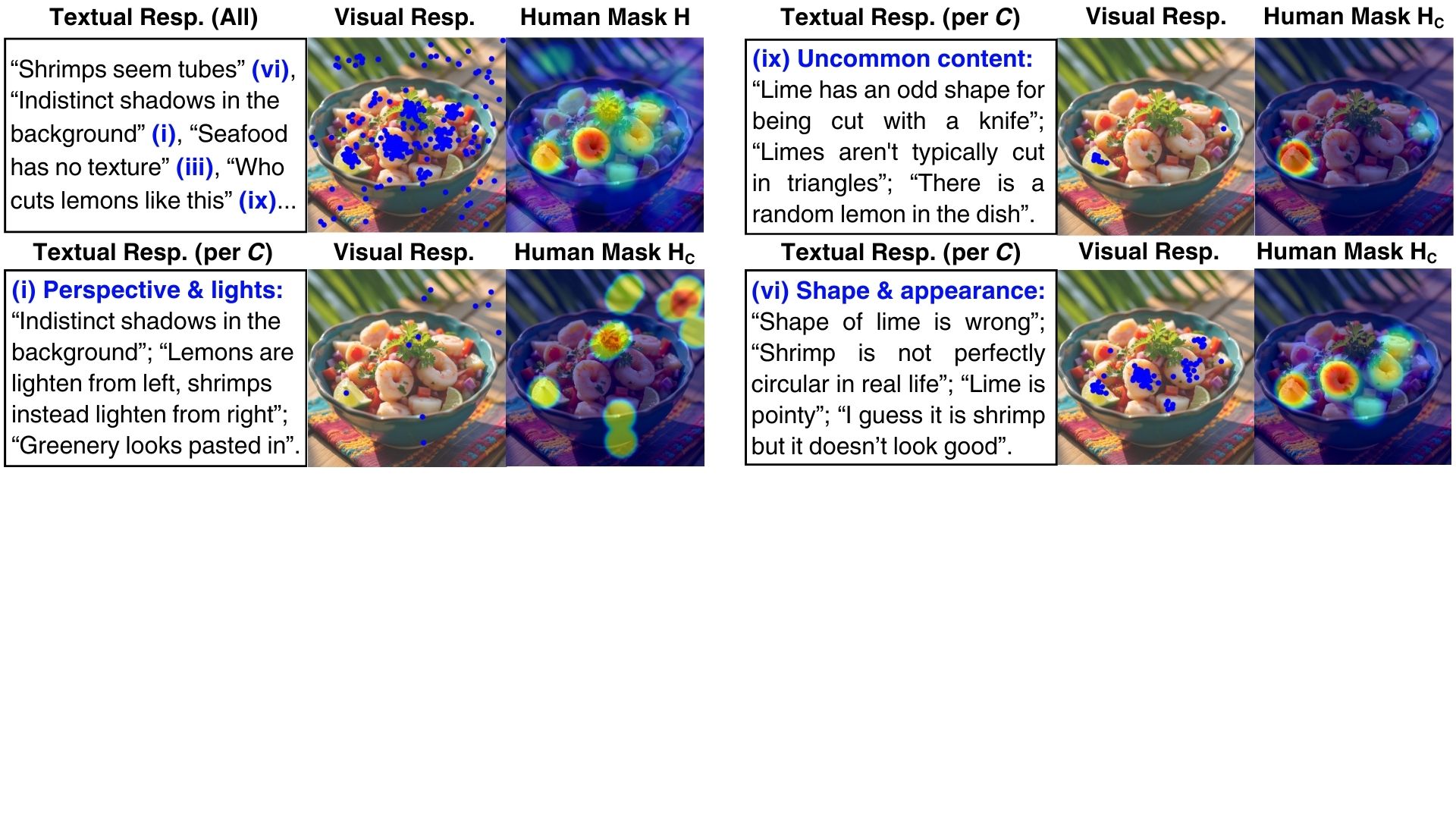}
	\caption{Example of text-based human masks $\mathbf{H}_{C}$ for different text categories $C$, indicated in blue. Again, $R$ is equal to $8.8\%$ of the image size, and $\alpha=3$.}
	\label{fig:txt}
\end{figure}

For each text category $C$, a text-based human mask $\mathbf{H}_{C}$ is generated as described in Section~\ref{sec:visual}, but considering only the participants' visual responses that are accompanied by a textual response falling within $C$ (See Figure~\ref{fig:txt}).

\begin{table*}[h]
	\centering
	\scriptsize
	\caption{Best average cosine similarity score between XAI masks and text-based human masks achieved across all XAI methods and AI detectors.}
	\vspace{0.2cm}
	\begin{minipage}[c]{0.48\textwidth} 
		\centering
		\begin{tabular}{|>{\arraybackslash}p{3.3cm}|>{\centering\arraybackslash}p{1.6cm}|}
			\hline
			\centering\textbf{Text category} & \textbf{Sim. Score} \\
			\hline
			(i) Perspective \& lights & 0.29\\ 
			(ii) Inconsistency & 0.25\\ 
			(iii) Texture \& details & 0.38\\ 
			(iv) Background & 0.18\\ 
			(v) Distortion \& blend & 0.32 \\
				\hline
			\cellcolor{gray!20}Visual quality (All) & \cellcolor{gray!20}0.47 \\
			\hline
		\end{tabular}
	\end{minipage}
	\hfill
	\begin{minipage}[c]{0.48\textwidth} 
		\centering
		\begin{tabular}{|>{\arraybackslash}p{3.3cm}|>{\centering\arraybackslash}p{1.6cm}|}
			\hline
			\centering\textbf{Text category} & \textbf{Sim. Score} \\
			\hline
			(vi) Shape \& appearance & 0.38 \\
			(vii) Synthetic perfection & 0.26 \\
			(viii) Absence of physics & 0.25 \\
			(ix) Uncommon content & 0.37 \\
			(x) Unknown design & 0.24\\
			(xi) VIP subject & 0.43\\
			(xii) Non-sense text & 0.22 \\
			\hline
			\cellcolor{gray!20}Realism of content (All) & \cellcolor{gray!20}0.46\\
			\hline
		\end{tabular}
	\end{minipage}

	\label{tab:sim-eval-}
\end{table*}

XAI methods achieve a significantly lower overall similarity to text-based human masks than Figure~\ref{fig:human_r_alpha}, suggesting AI detectors rely on a cohesive combination of low-level image features rather than isolating singular, complex semantic concepts.
As a result, training multi-modal AI detectors with textual explanations presents significant challenges, as the explanations are unlikely to align with human expectations given the functioning of current AI detection strategies. This finding supports the evaluation study of Large Multi-modal Models (LLMs) for AI detection proposed by Li~et al.~\cite{fakebench}, stressing the importance of incorporating more human-interpretable forensic elements in visual-language AI detection.

\section{Conclusion}
\label{sec:conclusions}
This work comprehensively evaluates 16 XAI methods for AI-generated image detection models, categorizing them by visual features and assessing them across detectors with varying architectures and training strategies. A novel survey-based evaluation, collecting both visual and textual responses from human participants, highlights significant alignment between human preferences and specific combinations of XAI methods and AI detectors, for certain image categories. The study reveals patterns and cues needed to enhance interpretability, which future scientific work should consider when developing explainable AI detectors.

\begin{credits}
\subsubsection{\ackname} This work was supported by the GADMO 2.0 project, co-funded by the European Union’s Digital Europe Programme (DIGITAL) and the Austrian Research Promotion Agency (FFG), under grant agreement No. 101083573, and by the defame Fakes project, funded by the Austrian security research programme KIRAS of the Federal Ministry of Finance (BMF).

\subsubsection{\discintname}
The authors have no competing interests to declare that are relevant to the content of this article.
\end{credits}

\bibliographystyle{splncs04}
\bibliography{egbib}

\end{document}